\documentclass[letterpaper]{article} 
\usepackage{aaai2026}  
\usepackage{times}  
\usepackage{helvet}  
\usepackage{courier}  
\usepackage[hyphens]{url}  
\usepackage{graphicx} 
\usepackage{natbib}  
\usepackage{caption} 
\usepackage{algorithm}
\usepackage{algorithmic}
\usepackage{booktabs}

\usepackage{newfloat}
\usepackage{listings}
\DeclareCaptionStyle{ruled}{labelfont=normalfont,labelsep=colon,strut=off} 
\floatstyle{ruled}
\newfloat{listing}{tb}{lst}{}
\floatname{listing}{Listing}
\nocopyright

\title{CBDES MoE: Hierarchically Decoupled Mixture-of-Experts for Functional Modules in Autonomous Driving}
\author {
    Qi Xiang\textsuperscript{\rm 1,2,3},
    Kunsong Shi\textsuperscript{\rm 1,2},
    Zhigui Lin\textsuperscript{\rm 2,4},
    Lei He\textsuperscript{\rm 1,2,}\thanks{Corresponding author: helei2023@tsinghua.edu.cn}
}
\affiliations {
    \textsuperscript{\rm 1}School of Vehicle and Mobility, Tsinghua University, Beijing 100084, China\\
    \textsuperscript{\rm 2}State Key Laboratory of Intelligent Green Vehicle and Mobility, Tsinghua University, Beijing 100084, China\\
    \textsuperscript{\rm 3}Department of Computing, Imperial College London, London SW7 2AZ, UK\\
    \textsuperscript{\rm 4}School of Automotive Engineering, Wuhan University of Technology, Wuhan 430070, China
}

\begin{document}

\maketitle

\begin{abstract}
Bird’s Eye View (BEV) perception systems based on multi-sensor feature fusion have become a fundamental cornerstone for end-to-end autonomous driving. However, existing multi-modal BEV methods commonly suffer from limited input adaptability, constrained modeling capacity, and suboptimal generalization. To address these challenges, we propose a hierarchically decoupled Mixture-of-Experts architecture at the functional module level, termed Computing Brain DEvelopment System Mixture-of-Experts (CBDES MoE). CBDES MoE integrates multiple structurally heterogeneous expert networks with a lightweight Self-Attention Router (SAR) gating mechanism, enabling dynamic expert path selection and sparse, input-aware efficient inference. To the best of our knowledge, this is the first modular Mixture-of-Experts framework constructed at the functional module granularity within the autonomous driving domain. Extensive evaluations on the real-world nuScenes dataset demonstrate that CBDES MoE consistently outperforms fixed single-expert baselines in 3D object detection. Compared to the strongest single-expert model, CBDES MoE achieves a 1.6-point increase in mAP and a 4.1-point improvement in NDS, demonstrating the effectiveness and practical advantages of the proposed approach.
\end{abstract}


\section{Introduction}

With the rapid development of autonomous driving technologies, 3D perception has become a cornerstone for building safe, intelligent, and reliable driving systems. Among the mainstream solutions, Bird's Eye View (BEV)-based multi-modal fusion frameworks—such as BEVFusion—have shown great promise by projecting raw sensor inputs from cameras and LiDAR into a unified spatial representation~\cite{huang2022bevdethighperformancemulticamera3d, li2022bevdepthacquisitionreliabledepth, li2022bevformerlearningbirdseyeviewrepresentation, liu2024bevfusionmultitaskmultisensorfusion, liang2022bevfusionsimplerobustlidarcamera}. These structured BEV features enable accurate 3D object detection and support a range of downstream driving tasks. Despite their success, most existing approaches adopt fixed single-backbone feature extractors for each modality. While this design choice simplifies the training pipeline, it severely restricts the model’s adaptability to diverse and dynamically changing driving environments, such as varying lighting, weather, and camera viewpoints. Moreover, the limited modeling capacity of single-backbone architectures compromises their ability to capture semantically rich and complex scene information, ultimately leading to degraded performance under domain shifts or task transitions.

To address these challenges, prior studies have explored adaptive modules such as dynamic convolutions and deformable attention~\cite{chen2020dynamicconvolutionattentionconvolution, xia2022visiontransformerdeformableattention, he2022masked}. While these techniques improve local flexibility, they remain confined within a rigid, monolithic network structure. Lacking coarse-grained architectural adaptability, such models fall short in dynamically reallocating capacity based on input variability—thereby hindering large-scale deployment in safety-critical autonomous driving systems. The Mixture-of-Experts (MoE) paradigm offers a compelling solution. By enabling dynamic expert selection through learned routing mechanisms, MoE models can flexibly balance computational efficiency with representational richness~\cite{6797059}. Although MoE has achieved remarkable success in natural language processing and vision-language tasks~\cite{lepikhin2020gshardscalinggiantmodels, fedus2022switchtransformersscalingtrillion, dai2024deepseekmoeultimateexpertspecialization, riquelme2021scalingvisionsparsemixture}, its potential in BEV-based 3D perception remains largely untapped. In particular, designing heterogeneous expert backbones and routing strategies tailored for multi-modal fusion poses unique challenges that have not yet been adequately addressed.

In this work, we propose CBDES MoE, a hierarchically decoupled Mixture-of-Experts architecture designed specifically for BEV perception tasks. Our method introduces a diverse pool of backbone experts, a lightweight gated routing network, and sparse activation mechanisms to enable efficient, adaptive, and scalable 3D perception. Comprehensive experiments on the nuScenes benchmark demonstrate that our approach consistently surpasses strong single-backbone baselines, especially under complex and diverse driving scenarios. Our main contributions are summarized as follows:

\begin{itemize}
    \item CBDES MoE introduces a novel expert-based architecture with a heterogeneous pool of multi-stage backbone designs. By supporting hierarchical decoupling and dynamic expert selection, the model achieves enhanced scene adaptability and feature representation capability.
    \item A lightweight, layered routing mechanism is developed, integrating convolutional operations, self-attention, and multi-layer perceptrons. This module enables end-to-end learning of the input-to-expert mapping, supporting sparse activation and efficient dynamic inference.
    \item Extensive experiments are conducted on the nuScenes dataset, demonstrating that CBDES MoE consistently surpasses strong single-backbone baselines, particularly under diverse and challenging environmental conditions.
\end{itemize}

\section{Related Work}
\subsection{Multi-Modal BEV Perception}
As autonomous driving systems evolve, multi-sensor information fusion has become a key technique for improving perception accuracy and robustness. Early approaches often relied on independent pipeline stages or point cloud-based fusion methods (e.g., PointPillars, VoxelNet), which struggled to leverage the rich texture and semantics of images~\cite{lang2019pointpillarsfastencodersobject, zhou2017voxelnetendtoendlearningpoint}. In recent years, BEV representations have gained popularity due to their unified spatial projection properties. Methods like BEVDet, BEVDepth, BEVFormer, and others have demonstrated significant performance gains in 3D detection and map segmentation by projecting multi-view camera images into the BEV space~\cite{huang2022bevdethighperformancemulticamera3d, li2022bevdepthacquisitionreliabledepth, li2022bevformerlearningbirdseyeviewrepresentation}.

Further, several works, such as BEVFusion, proposed feature-level multi-modal fusion frameworks that align and jointly model camera and LiDAR information within the BEV space~\cite{liu2024bevfusionmultitaskmultisensorfusion, liang2022bevfusionsimplerobustlidarcamera}. However, these methods typically adopt fixed, single-backbone architectures (e.g., ResNet, Swin Transformer), lacking the ability to adapt to diverse and dynamic input conditions.

\subsection{Dynamic Design Paradigms}
Autonomous driving scenarios exhibit high variability—lighting, weather, viewpoints, and road layouts—requiring perception systems to have flexible and powerful modeling capabilities. While single, fixed architectures perform reliably under controlled conditions, they often degrade under domain shifts or task transitions. Recent studies have introduced dynamic convolution and deformable attention to enable adaptive parameter adjustment based on input features, thereby improving robustness~\cite{chen2020dynamicconvolutionattentionconvolution, xia2022visiontransformerdeformableattention, he2022masked}.

Nevertheless, these efforts primarily operate at the fine-grained module level within a single architecture and do not offer macro-level architectural diversity or dynamic path scheduling. The need remains for a mechanism that enables both structural diversity and dynamic selection, while maintaining inference efficiency—an important challenge in autonomous perception research.

\subsection{Mixture-of-Experts (MoE) Architectures}
Originally proposed by Jacobs et al., the Mixture-of-Experts (MoE) architecture enables dynamic expert selection based on input-dependent gating, enhancing representational capacity while controlling computational cost~\cite{6797059}. In NLP, models such as GShard and Switch Transformers have demonstrated that sparsely activated experts can scale model capacity linearly without incurring heavy inference overhead~\cite{lepikhin2020gshardscalinggiantmodels, fedus2022switchtransformersscalingtrillion}. Recently, DeepSeekMoE further pushes this boundary by scaling MoE models to over 100 billion parameters while maintaining efficiency through expert sparsity and optimized routing~\cite{dai2024deepseekmoeultimateexpertspecialization}. In vision, Vision MoE has shown strong performance in tasks such as classification and detection by incorporating expert modules~\cite{riquelme2021scalingvisionsparsemixture}.

However, systematic MoE integration into autonomous driving's multi-modal BEV perception remains largely unexplored. Challenges include designing appropriate expert combinations, efficient gating, and maintaining cross-modal consistency in the BEV projection space.

\subsection{MoE Research in Autonomous Driving}
In the autonomous driving domain, pioneering works have begun incorporating MoE into end-to-end task learning. For example, ARTEMIS introduced MoE into trajectory planning, using dynamic routing to address performance degradation under ambiguous guidance conditions and achieving robust planning across scenarios~\cite{feng2025artemisautoregressiveendtoendtrajectory}. DriveMoE proposed a vision-based MoE for perception and a behavior-based MoE for decision-making, enabling disentangled multi-view processing and diverse driving skills~\cite{yang2025drivemoemixtureofexpertsvisionlanguageactionmodel}. It achieved state-of-the-art results on the Bench2Drive benchmark.

These works confirm that MoE can enhance diversity and adaptability in end-to-end planning and decision-making modules. Nonetheless, the above studies primarily focus on end-to-end planning and decision layers. There is a lack of systematic solutions for implementing hierarchical, decoupled dynamic expert selection within multi-modal BEV perception systems.

\begin{figure*}[t]
\centering
\includegraphics[width=0.98\textwidth]{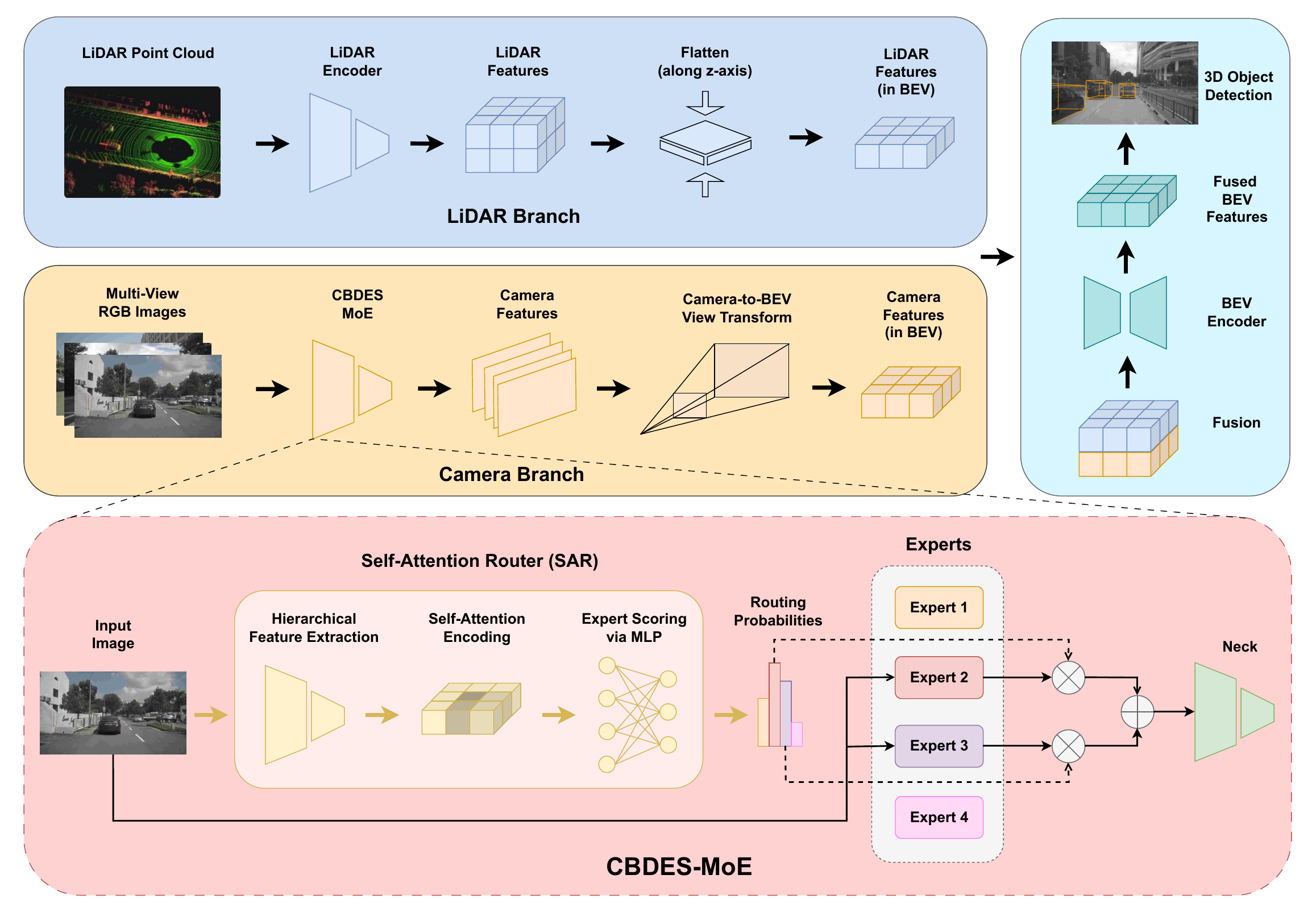}
\caption{Overview of the proposed CBDES MoE architecture. CBDES MoE employs a Self-Attention Router (SAR) to extract global features from the input image and generate expert routing probabilities. Several structurally diverse experts receive the shared input, and the top-k experts are selected per image based on the router’s output. The selected experts' features are fused and passed to the neck module for downstream tasks. The SAR enables dynamic, image-level expert selection with minimal overhead, enhancing both adaptability and efficiency.}
\label{fig:arch}
\end{figure*}

\section{Method}
In this section, we present the architecture and design details of our proposed CBDES MoE, a novel multi-expert design for multi-modal BEV perception tasks in autonomous driving. The CBDES MoE module is introduced as a plug-and-play backbone to enhance feature representation by leveraging architectural diversity and input-adaptive routing. Our design specifically addresses the limitations of single-backbone fusion networks by enabling heterogeneous expert composition, input-dependent dynamic routing, and feature-level aggregation in the BEV space.

\subsection{Overall Framework}
The CBDES MoE module is integrated within a BEVFusion-style framework, where multi-camera images are projected into the BEV space and fused with LiDAR/radar modalities for 3D perception tasks such as object detection and map segmentation. We replace the conventional static image backbone with a heterogeneous mixture-of-experts module to extract image features prior to BEV projection. Each expert processes input images independently before projecting features to the BEV space using camera-to-BEV view transformation layers. All subsequent operations such as modality fusion and task heads remain compatible with standard pipelines like BEVFusion.

As shown in Figure~\ref{fig:arch}, the CBDES MoE consists of four architecturally diverse expert networks, a lightweight learnable routing module, and a soft feature fusion mechanism for expert outputs. Each component is designed with efficiency, diversity, and input-adaptivity in mind, enabling the model to selectively utilize expert pathways under different visual conditions and task demands.

\subsection{Heterogeneous Expert Set Design}

A major innovation of CBDES MoE lies in the explicit architectural heterogeneity among the expert backbones. Instead of using experts of the same architecture, we design an expert pool composed of four fundamentally distinct vision backbones, each representing a different paradigm in deep visual representation learning.

\begin{itemize}
    \item \textbf{Swin Transformer}: A hierarchical transformer that captures long-range dependencies with window-based self-attention and shifted window mechanisms~\cite{liu2021swintransformerhierarchicalvision}. It is especially effective for capturing global spatial structures in scenes with large field-of-view coverage.
    \item \textbf{ResNet}: A classical convolutional backbone using residual connections~\cite{he2015deepresiduallearningimage}. Despite its simplicity, it provides strong inductive biases for local structure encoding and edge detection, particularly helpful under poor lighting or texture-less conditions.
    \item \textbf{ConvNeXt}: A modernized convolutional network architecture inspired by transformers but retaining full-convolutional design~\cite{liu2022convnet2020s}. It strikes a balance between locality and scalability, offering powerful representations with fewer handcrafted priors.
    \item \textbf{Pyramid Vision Transformer (PVT)}: A hierarchical transformer designed for dense prediction tasks~\cite{wang2021pyramidvisiontransformerversatile}. It introduces spatial reduction attention to balance global context modeling and computational cost, making it suitable for multi-scale object modeling.
\end{itemize}

By integrating four structurally diverse expert backbones, the model benefits from varied inductive biases and representational capabilities. These experts differ not only in architecture but also in their receptive field characteristics, parameterization patterns, and computational footprints. Such heterogeneity is crucial for modeling the vast variability found in real-world autonomous driving scenes (e.g., urban vs. rural, day vs. night, clear vs. foggy). Each expert excels in capturing distinct image patterns, making the system more resilient to dynamic and diverse road scenes.

\subsection{Self-Attention Router (SAR)}

To dynamically assign the most appropriate expert to each input image, we propose a Self-Attention Router (SAR) that leverages convolutional feature extraction followed by a lightweight self-attention mechanism and MLP classifier.

The SAR takes as input a feature map $\mathbf{X} \in B \times C \times H \times W$, where $B$ is the batch size, $C$ is the number of input channels, and $H \times W$ is the spatial resolution. The module then proceeds through three main stages: Hierarchical feature extraction, self-attention encoding, and expert scoring via MLP.

\paragraph{Hierarchical Feature Extraction.}
We begin with a sequence of convolutional and pooling layers that reduce the spatial resolution while increasing channel dimensionality:

\begin{equation}
    \mathbf{X}_1 = \mathrm{Pool}_1(\mathrm{Conv}_{3 \times 3}^{32}(\mathbf{X}))
\end{equation}
\begin{equation}
    \mathbf{X}_2 = \mathrm{Pool}_2(\mathrm{Conv}_{3 \times 3}^{64}(\mathbf{X}_1))
\end{equation}
\begin{equation}
    \mathbf{X}_3 = \mathrm{Pool}_3(\mathrm{Conv}_{3 \times 3}^{d_{emb}}(\mathbf{X}_2))
\end{equation}

where $d_{emb}$ is the attention embedding dimension (e.g., 128). Each ConvModule contains a convolutional layer followed by batch normalization and PReLU activation. Pooling is applied using $2 \times 2$ max pooling with stride 2.

\paragraph{Self-Attention Encoding.}

The result of the previous stage $\mathbf{X}_3 \in \mathcal{R}^{B \times d_{emb} \times H' \times W'}$ is then reshaped into a token sequence:

\begin{equation}
    \mathbf{T} = Flatten(\mathbf{X}_3) \in \mathcal{R}^{B \times N \times d_{emb}}, \quad N = H' \cdot W'
\end{equation}

To model the global interactions among spatial tokens, we apply a multi-head self-attention (MHA) layer:

\begin{equation}
    \mathbf{T'} = LayerNorm(MHA(\mathbf{T}))
\end{equation}

This operation enables the router to integrate spatial dependencies and learn a richer representation of the global scene context.

The output sequence $\mathbf{T'}$ is then averaged over the token dimension to produce an image-level embedding:

\begin{equation}
    \mathbf{G} = \frac{1}{N}\sum_{i=1}^N\mathbf{T}_i' \in \mathcal{R}^{B \times d_{emb}}
\end{equation}

\paragraph{Expert Scoring via MLP.}

The global descriptor $\mathbf{G}$ from the previous stage is passed through a 3-layer MLP with PReLU activations to produce the expert logits:

\begin{equation}
    \mathbf{S} = MLP(\mathbf{G}) \in \mathcal{R}^{B \times K}
\end{equation}

where $K$ is the number of experts. Finally, the router applies a softmax to obtain the routing probabilities:

\begin{equation}
    \mathbf{P} = softmax(\mathbf{S})
\end{equation}

Each row $\mathbf{P}_i \in \mathcal{R}^K$ represents the image-level soft assignment to the $K$ available experts.

\bigskip
The image-level routing mechanism enables the model to adapt its computation path based on input semantics. For instance, rainy scenes may be routed to transformer-heavy experts, while texture-rich urban scenes may benefit from convolutional networks. This adaptability improves the model’s robustness to domain shifts and rare or complex scenes, addressing a common shortcoming of static models.

The router itself is lightweight, consisting of a few convolutional layers, a single-layer multi-head self-attention mechanism, and a compact MLP. Despite its simplicity, it can effectively summarize global scene-level information and produce semantically meaningful routing scores. It enables the system to dynamically assign different weights to each expert per input image. This routing flexibility empowers the model to automatically specialize each expert for different types of scenes, lighting conditions, or spatial layouts without manual intervention.

\subsection{Expert Feature Extraction}
Each expert receives the image as input and produces its own processed feature:

\begin{equation}
    F^{(k)} = \mathcal{E}_k(F_{img}), \quad k \in 1, 2, 3, 4
\end{equation}

where $\mathcal{E}_k$ denotes the $k$-th expert network.

Notably, all expert outputs maintain the same spatial resolution and channel dimension (via adapter layers if needed) to ensure compatibility during fusion. The architectural difference leads each expert to specialize in different visual patterns—some experts may excel in detecting long-range vehicles, while others may better segment road boundaries.

\subsection{Soft Weighted Feature Fusion}
After obtaining the routing scores and expert outputs, we perform a weighted soft fusion across experts using the routing maps. The final fused feature map $F_{fused} \in \mathcal{R}^{B \times C \times H \times W}$ is computed as:

\begin{equation}
    F_{fused}^{b, c} = \sum_{k=1}^K r_{b,k} \cdot F_{b,c}^{(k)}
\end{equation}

This mechanism enables smooth transitions between experts and avoids hard routing instability, while still maintaining sparse activation patterns due to softmax sharpening during training.

\subsection{Efficient Inference via Sparse Expert Activation}

At inference time, we activate only the top-1 expert per image based on the router's output. This sparse expert activation strategy greatly reduces computational cost compared to evaluating all experts, while maintaining competitive accuracy due to the router's discriminative capacity.

Formally, for an input batch of size $B$, only $B$ forward passes through expert backbones are required, rather than $B \times K$, where $K$ is the number of experts. This yields a linear scaling with respect to batch size and makes the system feasible for real-time applications on edge hardware.

\subsection{Load Balancing Regularization}
One of the common issue for MoE is the expert collapse problem, where the router consistently selects only a subset of experts, leaving others underutilized. This imbalance not only wastes model capacity but also reduces the benefits of expert diversity. To address this, we introduce a load balancing regularization term that encourages uniform usage of all experts across the dataset.

Let $\mathbf{P} \in \mathcal{R}^{N \times K}$ denote the routing probability matrix, where $N$ is the number of samples (e.g., images in a batch), and $K$ is the number of experts. Each row $\mathbf{P}_i \in \mathcal{R}^k$ represents the soft routing probabilities assigned to the $K$ experts for the $i$-th sample, such that $\sum_{j=1}^{K} \mathbf{P}_{i,j} = 1$.

The expert mean activation is defined as:

\begin{equation}
    \bar{\mathbf{p}} = \frac{1}{N}\sum_{i=1}^N \mathbf{P}_i \in \mathcal{R}^k
\end{equation}

and the total routing load per expert as:

\begin{equation}
    \mathbf{l} = \sum_{i=1}^N \mathbf{P}_i \in \mathcal{R}^k
\end{equation}

Then, the load balance loss is defined as:

\begin{equation}
    \mathcal{L}_{balance} = \sum_{j=1}^K \bar{p}_j \cdot l_j
\end{equation}

This formulation penalizes the joint deviation between expert usage frequency ($\bar{p}_j$) and the accumulated load ($l_j$). It reaches its minimum when all experts are assigned uniformly across the batch, thus promoting equitable participation of all experts in training.

This regularization term is integrated into the overall training objective:

\begin{equation}
    \mathcal{L} = \mathcal{L}_{task} + \lambda\mathcal{L}_{balance}
\end{equation}

where $\lambda$ is a hyperparameter controlling the trade-off between task performance and routing diversity. In practice, a small $\lambda$ (e.g., $0.01$) is sufficient to encourage load balancing without disrupting task convergence.

This load balancing regularization ensures that all experts are engaged during training, leading to richer specialization and avoiding export under-utilization.

\subsection{Integration with BEVFusion}
The fused feature map $F_{fused}$ is passed through the standard camera-to-BEV projection module in BEVFusion. This module uses camera intrinsic and extrinsic matrices to backproject 2D features into a common BEV grid. The resulting BEV feature is then fused with other modalities (e.g., LiDAR, radar) and passed to task-specific heads for 3D object detection, semantic segmentation, or instance segmentation.

Due to the plug-and-play nature of our design, CBDES MoE can be seamlessly integrated into various BEV-based perception frameworks without requiring changes to projection logic or downstream heads.

\subsection{Training Strategy}
We train CBDES MoE end-to-end using the same loss functions as standard BEVFusion pipelines (e.g., focal loss for detection), with the addition of expert load balancing loss to equalize expert usage and promote specialization.

The Self-Attention Router is trained jointly with the rest of the model. During training, we use soft gating (i.e., weighted sum of all expert outputs) to ensure differentiability. At inference, we switch to \textbf{top-1 expert activation} for each image to reduce computational overhead. Unless otherwise specified, all results are reported under this inference regime.

We use mixed precision training to reduce training time and GPU memory usage. All experts are optimized jointly, with routing parameters trained via standard backpropagation.

\section{Experiments}

To evaluate the effectiveness of the proposed \textbf{CBDES MoE} framework, we conduct comprehensive experiments focusing on 3D object detection in autonomous driving scenarios. The goal is to assess both the performance gains brought by dynamic heterogeneous expert selection and the benefits of load balance regularization. We compare CBDES MoE with several strong single-expert baselines and conduct ablation studies to isolate the contribution of each component.

\begin{figure*}[t]
\centering
\includegraphics[width=0.92\textwidth]{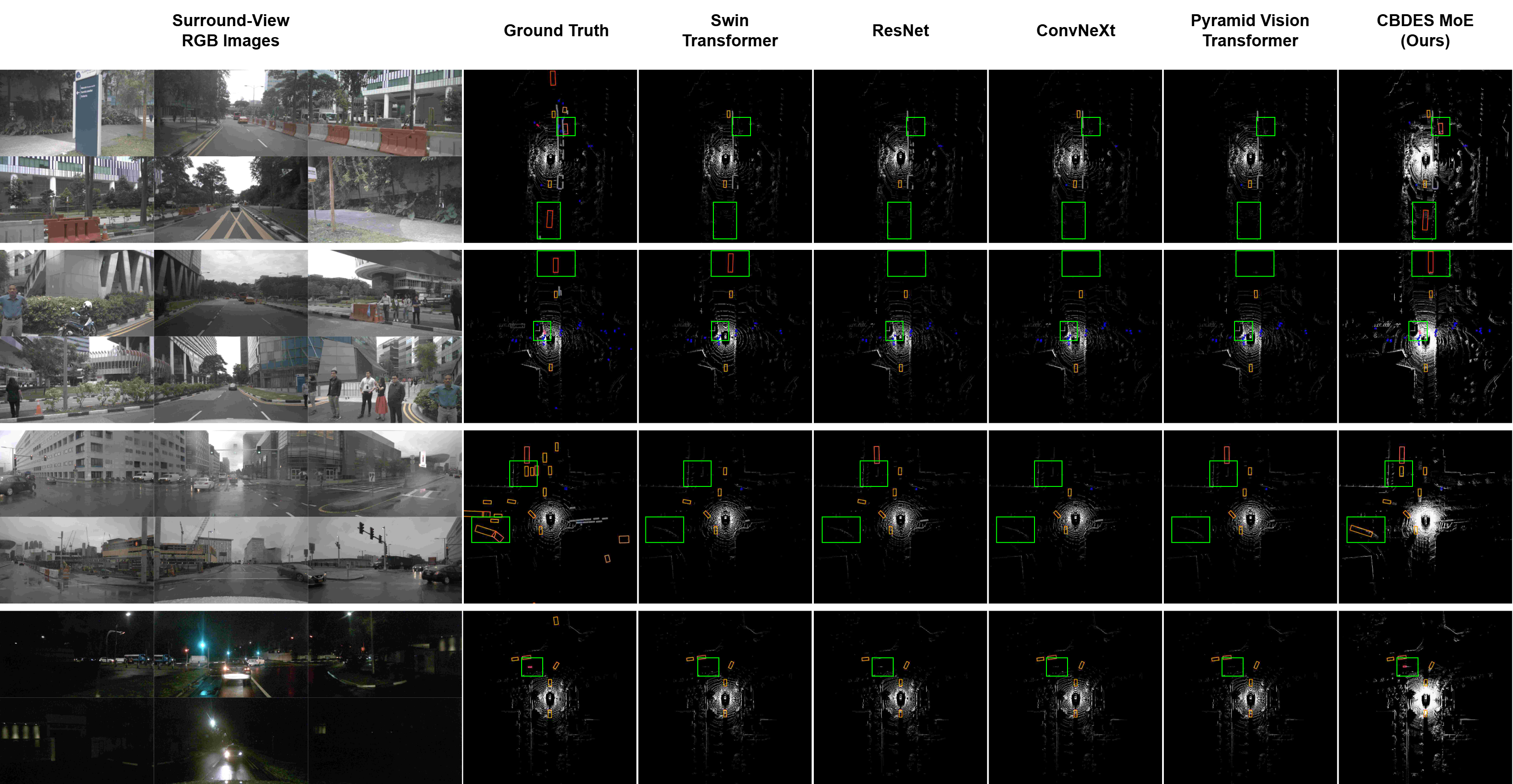}
\caption{Bird’s-Eye View (BEV) 3D object detection results. From left to right: (1) Surround-view RGB images from six cameras; (2) Ground truth labels (yellow: cars, blue: pedestrians, red: bicycles/motorcycles/buses, gray: barriers); (3)–(7) Predictions from using Swin Transformer, ResNet, ConvNeXt, PVT backbone, and our model, respectively. Green rectangles highlight key performance differences between single-expert models and our model, particularly in regions where our approach better aligns with the ground truth.}
\label{fig:visualization}
\end{figure*}

\subsection{Experimental Setup}

We evaluate the performance of our CBDES MoE by incorporating it into the official BEVFusion codebase. The four experts—Swin Transformer, ResNet, ConvNeXt, and PVT—are all pretrained on ImageNet-1K and adapted to match the input-output interfaces of the BEVFusion camera branch.

\paragraph{Dataset.}
We perform all experiments on the widely-used \textbf{nuScenes} dataset~\cite{caesar2020nuscenesmultimodaldatasetautonomous}, a large-scale autonomous driving benchmark that contains over 1,000 scenes with synchronized multi-view camera and LiDAR data. Following the official split, we use 700 scenes for training, 150 for validation, and 150 for testing. The benchmark provides 10 categories for 3D object detection, including car, truck, bus, trailer, construction vehicle, pedestrian, motorcycle, bicycle, traffic cone, and barrier.

\paragraph{Evaluation Criteria.}
Our main task is \textbf{3D object detection}, and we adopt two standard evaluation metrics from the nuScenes benchmark: \textbf{mean Average Precision (mAP)}, computed using a 2D center distance threshold in the BEV plane, and \textbf{nuScenes Detection Score (NDS)}, a composite metric that considers mAP along with true positive translation, scale, orientation, velocity, and attribute errors.

\paragraph{Computational Resources.}
All experiments are conducted on a server with an Intel(R) Xeon(R) Gold 5318Y processor and 2$\times$ NVIDIA A100 GPUs (80GB memory each). We used Python 3.8 to execute the experiments. Mixed precision training (FP16) is used to accelerate training and reduce memory consumption. Distributed data-parallel training is implemented using PyTorch and MMCV.

\paragraph{Hyper-parameters.}
During training, we use the following settings across all experiments: For optimizer, we use AdamW with initial learning rate 5e-5 and weight decay 0.01. For learning rate config, we use cosine annealing with warm-up of 500 iterations. The batch size is set to 4 samples per GPU, and we run the experiments for 6 epochs. We use the same data augmentation and depth estimation strategies as BEVFusion for fair comparison.

\subsection{Comparison with Single-Expert Models}
To evaluate the contribution of expert diversity and dynamic routing, we compare \textbf{CBDES MoE} with four baselines, each using a single expert backbone: Swin Transformer, ResNet, ConvNeXt, and PVT. All models share the same BEV fusion, detection head, and training configuration to ensure fair comparison. The single-expert variants omit the MoE structure and routing mechanism.

\begin{figure}[!t]
\centering
\includegraphics[width=0.95\columnwidth]{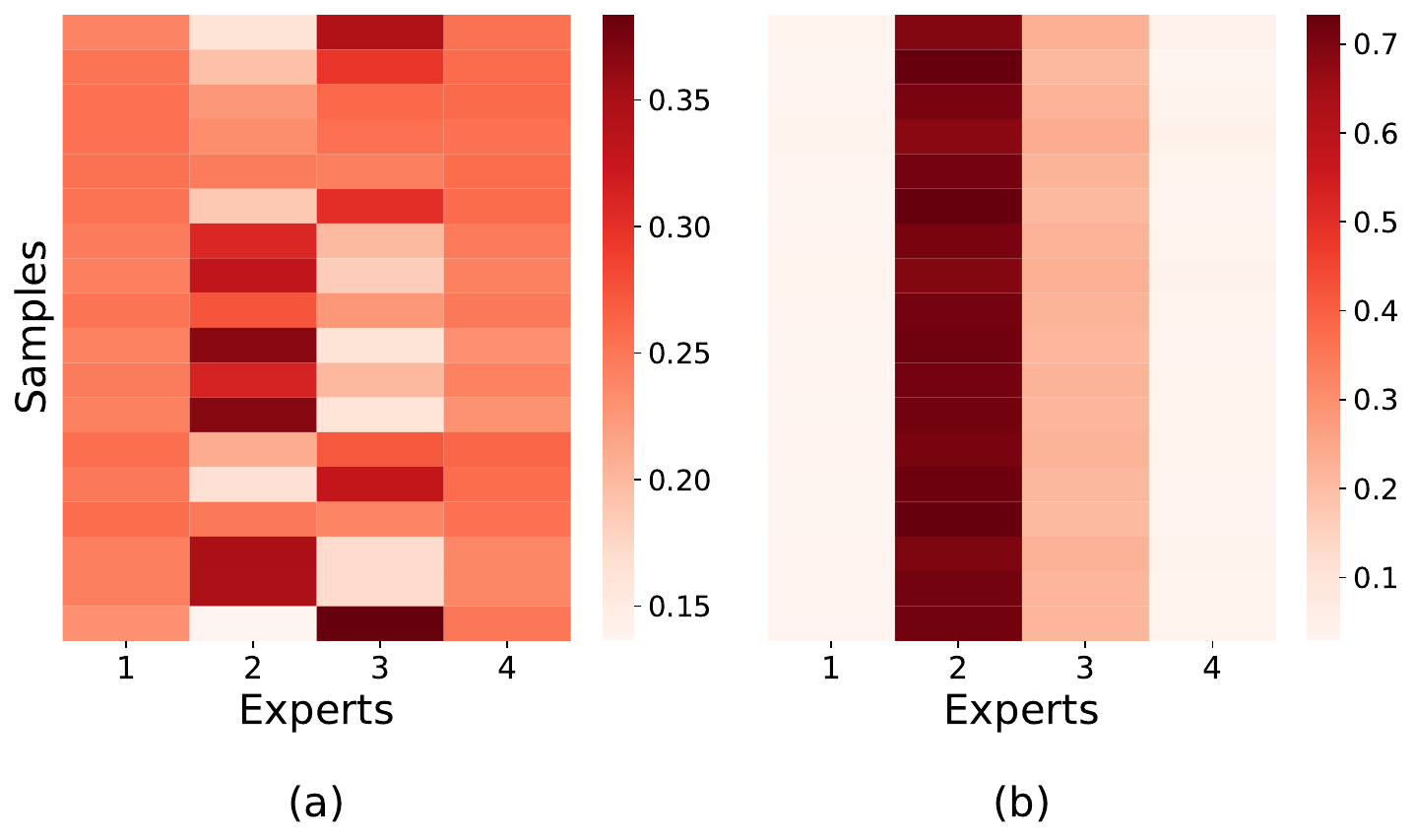}
\caption{Visualization of routing probabilities with (a) and without (b) load balance regularization. The inclusion of load balance loss results in more balanced expert utilization and avoids expert collapse. Each row represents a sample, and each column corresponds to one of the four experts.}
\label{fig:routing_probs}
\end{figure}

\begin{figure}[t]
\centering
\includegraphics[width=0.95\columnwidth]{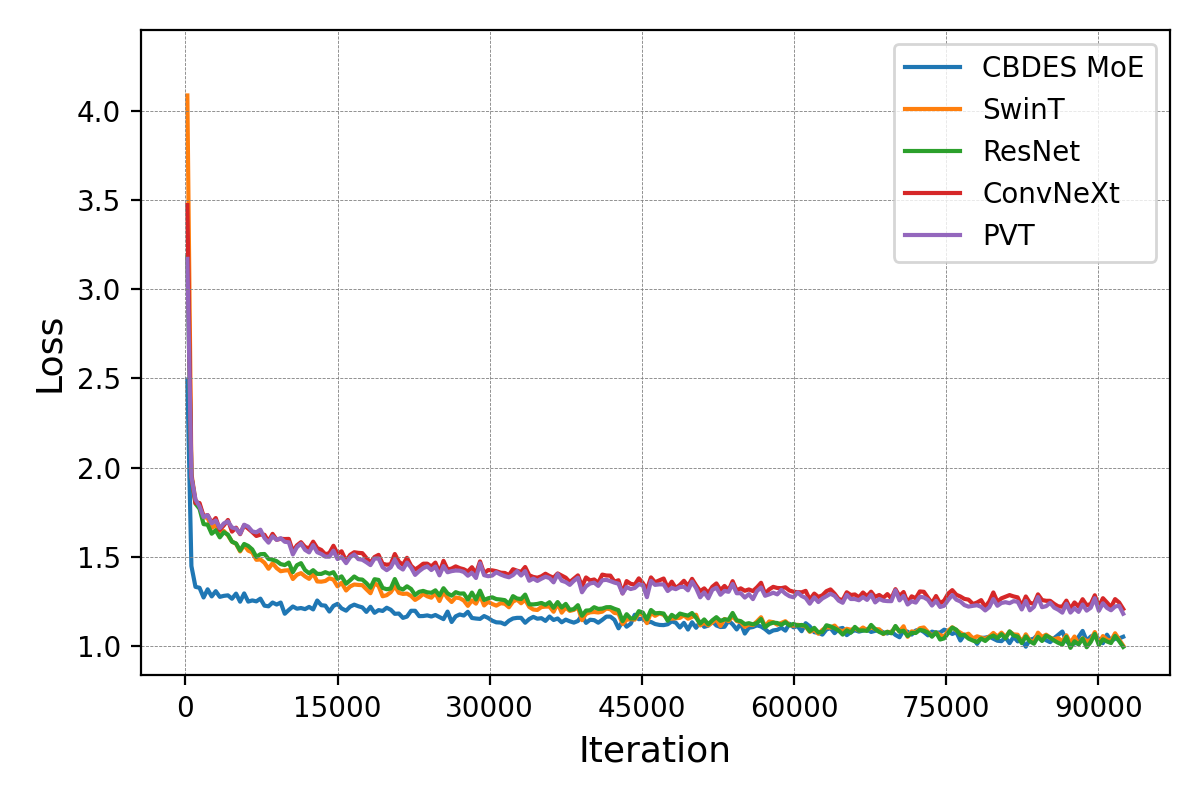}
\caption{Training loss curves of CBDES MoE and four single-expert baselines.}
\label{fig:training_progress}
\end{figure}

\begin{table}[ht]
\centering
\begin{tabular}{lccc}
\toprule
\textbf{Model} & \textbf{mAP}$\uparrow$ & \textbf{NDS}$\uparrow$\\
\midrule
BEVFusion-Swin Transformer & 64.0            & 65.6 \\
BEVFusion-ResNet       & 63.3              & 65.2 \\
BEVFusion-ConvNeXt       & 61.6              & 65.2 \\
BEVFusion-PVT            & 62.4              & 65.7 \\
\midrule
\textbf{CBDES MoE (Ours)} & \textbf{65.6} & \textbf{69.8} \\
\bottomrule
\end{tabular}
\caption{Performance comparison between single-expert models and \textbf{CBDES MoE}.}
\label{tab:moe_vs_single}
\end{table}

As shown in Table~\ref{tab:moe_vs_single}, CBDES MoE achieves consistently higher mAP and NDS compared to all four single-expert baselines. This validates the effectiveness of expert diversity and the adaptive routing mechanism in improving 3D detection performance.

In addition to accuracy metrics, we also compare training progress in Figure~\ref{fig:training_progress}. CBDES MoE exhibits faster convergence and consistently lower loss throughout training, indicating improved optimization stability and learning efficiency.

To further illustrate these quantitative gains, Figure~\ref{fig:visualization} presents qualitative comparisons of 3D object detection results in BEV  across four representative scenes under varying environmental conditions. The first two rows depict normal daytime scenarios with clear visibility, while the third and fourth rows show more challenging conditions: the third features rain and fog, and the fourth is captured during nighttime. Across all scenes, CBDES MoE produces results that more closely align with the ground truth, especially in regions highlighted by green boxes.

In clear daytime scenes, CBDES MoE demonstrates stronger consistency and fewer missed detections compared to single-expert models. Under adverse conditions—such as poor visibility due to fog or limited illumination at night—our model outperforms the baselines by maintaining robust detections where others struggle with false positives or complete failure to localize objects. These results underscore the strength of dynamic expert routing: by selecting the most appropriate expert per input, CBDES MoE adapts better to environmental variation and demonstrates superior generalization across diverse real-world conditions.

\subsection{Effect of Load Balance Regularization}
We further analyze the impact of the load balance regularization on detection performance. Two variants of CBDES MoE are compared: one trained with the load balance loss, and one without it.

\begin{table}[ht]
\centering
\begin{tabular}{lccc}
\toprule
\textbf{Model} & \textbf{mAP}$\uparrow$ & \textbf{NDS}$\uparrow$ \\
\midrule
CBDES MoE (w/o LB) & 63.4 & 65.8 \\
\textbf{CBDES MoE (with LB)} & \textbf{65.6} & \textbf{69.8} \\
\bottomrule
\end{tabular}
\caption{Impact of load balance regularization on performance.}
\label{tab:lb_ablation}
\end{table}

To better understand the effect of the load balance loss on expert utilization, we visualize the router's behavior in Figure~\ref{fig:routing_probs}. The two heatmaps show the soft routing probabilities assigned to each expert for both settings: with and without the load balance regularization term.

Without the load balancing loss (b), the routing probabilities are highly skewed toward one dominant expert, leading to severe expert imbalance and underutilization of the remaining experts. In contrast, when load balance loss is applied (a), the routing probabilities are more evenly distributed across all four experts, and the final selections exhibit greater diversity. This confirms that the regularization term encourages the router to explore the full expert space and prevents collapse into degenerate routing patterns.

The improved expert diversity directly contributes to better performance, as demonstrated in Table~\ref{tab:lb_ablation}, where both mAP and NDS are higher under balanced expert usage. These results highlight the necessity of incorporating load balancing in MoE-based perception systems to fully exploit their modeling capacity.

\section{Conclusion, Limitations and Future Work}
In this paper, we propose CBDES MoE, a novel hierarchically decoupled MoE framework tailored for functional modules in autonomous driving. By integrating four structurally diverse expert backbones—Swin Transformer, ResNet, ConvNeXt, and Pyramid Vision Transformer—and introducing a lightweight Self-Attention Router for dynamic expert selection at the image level, our model adaptively activates the most suitable expert for each input. A load balance regularization term is further introduced to prevent expert collapse and ensure stable training. Experimental results on the nuScenes dataset demonstrate that CBDES MoE consistently outperforms single-expert baselines in terms of mAP and NDS, confirming the effectiveness of the proposed framework.

\textbf{Limitations and future work:} Despite its strong performance, CBDES MoE has several limitations. Currently, expert routing is performed only at the image level; future work could explore patch-wise or region-aware routing for finer-grained adaptation. Extending our method to handle multi-task setups (e.g., segmentation and tracking) or incorporating cross-modal routing signals from LiDAR could further enhance generalization. We also plan to investigate automated expert architecture search and hardware-aware model compression to optimize scalability and deployment.

\section{Acknowledgments}

This work was supported by the National Key R\&D Program of China, Project “Development of Large Model Technology and Scenario Library Construction for Autonomous Driving Data Closed-Loop” (Grant No. 2024YFB2505501), and the Guangxi Key Scientific and Technological Project, Project “Research and Industrialization of High-Performance and Cost-Effective Urban Pilot Driving Technologies” (Grant No. Guangxi Science and Technology AA24206054).

\bibliography{CBDES_MoE}


\end{document}